\pdfoutput=1

\documentclass[11pt]{article}

\usepackage{acl}

\usepackage{times}
\usepackage{latexsym}

\usepackage[T1]{fontenc}

\usepackage[utf8]{inputenc}

\usepackage{microtype}

\usepackage{booktabs}
\usepackage{todonotes}

\title{Sample, Translate, Recombine: Leveraging Audio Alignments for Data Augmentation in End-to-end Speech Translation}

\author{Tsz Kin Lam$^{1}$ \and Shigehiko Schamoni$^{2,1}$ \and Stefan Riezler$^{1,2}$ \\
  $^{1}$Department of Computational Linguistics, Heidelberg University \\
  $^{2}$Interdisciplinary Center for Scientific Computing (IWR), Heidelberg University \\
  \texttt{\{lam,schamoni,riezler\}@cl.uni-heidelberg.de} \\
}

\begin{document}

\maketitle

\begin{abstract}

End-to-end speech translation relies on data that pair source-language speech inputs with corresponding translations into a target language. Such data are notoriously scarce, making synthetic data augmentation by back-translation or knowledge distillation a necessary ingredient of end-to-end training. In this paper, we present a novel approach to data augmentation 
that leverages audio alignments, linguistic properties, and translation. 
First, we augment a transcription by \textit{sampling} from a suffix memory that stores text and audio data. Second, we \textit{translate} the augmented transcript. Finally, we \textit{recombine} concatenated audio segments and the generated translation. Besides training an MT-system, we only use basic off-the-shelf components without fine-tuning. 
While having similar resource demands as knowledge distillation, adding our method delivers consistent improvements of up to 0.9 and 1.1 BLEU points on five language pairs on CoVoST\,2 and on two language pairs on Europarl-ST, respectively.

\end{abstract}

\section{Introduction}

End-to-end automatic speech translation (AST) relies on data that consist only of speech inputs and corresponding translations. Such data are notoriously limited. Data augmentation approaches attempt to compensate the scarcity of such  data by generating synthetic data by translating transcripts into foreign languages or by back-translating target-language data via text-to-speech synthesis (TTS) \citep{PinoETAL:19,JiaETAL:19}, or by performing knowledge distillation using a translation system trained on gold standard transcripts and reference translations \citep{InagumaETAL:21}. In this paper, we present a simple, resource conserving approach that does not require TTS and yields improvements complementary to knowledge distillation (KD). 

For training cascaded systems, monolingual data for automatic speech recognition and textual translation data for machine translation can be used, reducing the problem of scarcity. Cascaded systems, however, suffer from error propagation, which has been addressed by using more complex intermediate representations such as $n$-best machine translation (MT) outputs or lattices \cite[\textit{inter alia}]{BertoldiFederico05,BeckETAL19} or by modifying training data to incorporate errors from automatic speech recognition (ASR) and MT \cite{RuizETAL:15,LamETAL:21a}. End-to-end systems are unaffected by this kind of error propagation and are able to surpass cascaded systems if trained on sufficient amounts of data \cite{SperberPaulik20}.

Our approach transfers an idea on aligned data augmentation that has been presented for ASR \citep{LamETAL:21} to aligned data augmentation in AST. Similar to aligned data augmentation for ASR, we utilize forced alignment information to create unseen training pairs in a structured manner. 
Our augmentation procedure consists of the following steps: (1)
\textit{Sampling} of a replacement suffix of a transcription and its aligned speech representations, guided by linguistic constraints. (2) \textit{Translation} of the transcription containing the new suffix. 
(3) \textit{Recombination} of audio data containing the new suffix and the generated translation to distill a new training pair. We thus use the acronym STR (Sample, Translate, Recombine) to refer to our method.

In comparison to \citet{PinoETAL:19} and \citet{JiaETAL:19} who use TTS to generate synthetic speech, we create new examples by recombining real human speech. This reduces the problem of overfitting to synthetic data as for example in SkinAugment \cite{McCarthyETAl20} where synthetic audio is generated by auto-encoding speaker conversions.
The basic idea of our method is comparable to data augmentation techniques for images such as CutMix \cite{yunetal-2019-cutmix} where images are blended together to form new data examples. However, CutMix selects images randomly, while we recombine phrases in a structured manner.

Our experimental evaluation is conducted for five language pairs on the CoVoST\,2 dataset \cite{wang2020covost} and for two language pairs on the Europarl-ST \cite{iranzo2020europarl} dataset. We find considerable improvements for all language pairs on all datasets for our approach on top of KD. Our approach can be seen as an enhancement of \citet{InagumaETAL:21}'s KD approach since it requires roughly the same computational resources and consistently improves their gains.

\begin{figure*}
\centering
\includegraphics[width=.94\linewidth]{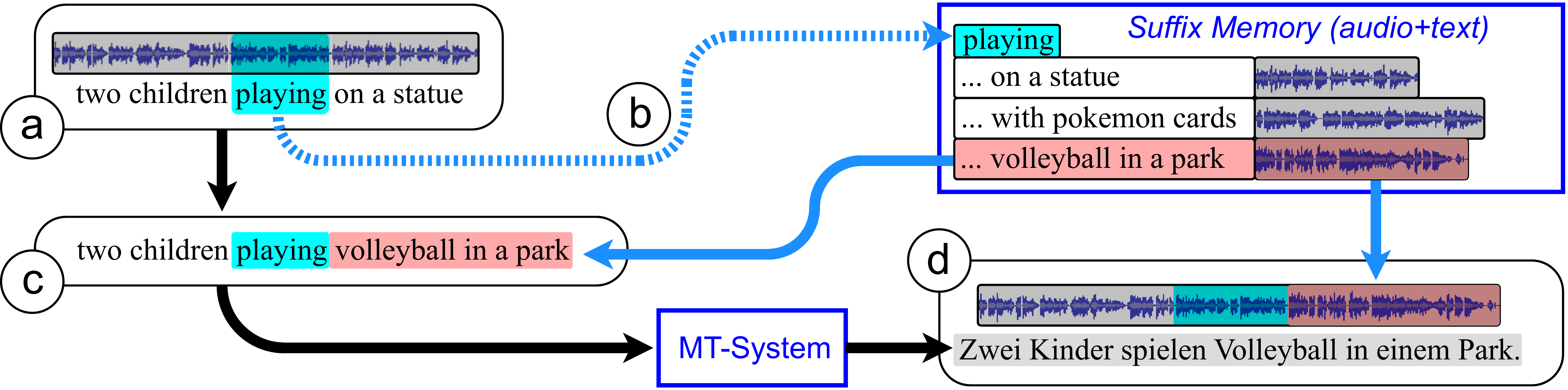}
\caption{(a) Select a pivoting token, e.g., ``playing''. (b) Retrieve suitable text-audio entries from the suffix memory to sample a replacement. (c) Compile a new transcription containing prefix, pivoting token, and replacement suffix. (d) Recombine a new training example by translating the new transcription and concatenating the audio sections.}
\label{fig:procedure}
\vspace{-2mm}
\end{figure*}

\section{Method}

Our method exploits audio-transcription alignment information to generate previously unseen data pairs for end-to-end AST training. 
By applying a Part-of-Speech (POS) Tagger on a sentence, we identify potential ``pivoting tokens'' where the token's prefix or suffix, i.e., the preceding or succeeding tokens, can be exchanged between other sentences containing the same token of the same syntactic function. We then sample possible suffixes for that token from a suffix memory containing text and audio suffixes, and concatenate the prefix, verb, and suffix to generate a new transcription. 
Then, an MT system translates the new transcription, picking up on the idea of knowledge distillation in AST \cite{InagumaETAL:21}. The MT system is trained or fine-tuned on the transcription-translation pairs. 
Finally, using the previously sampled audio suffix, we concatenate prefix, verb, and suffix audio together with the MT generated translation to recombine a new audio-translation pair for end-to-end AST training. 

Our augmentation method implements linguistic constraints by making use of the transcription's syntactic structure in combination with alignment information. Effectively, we exploit the strict SVO-scheme of English sentences as we select the verb as our pivoting token. Our method is applicable to other languages, however, it will require more effort to identify exchangeable syntactic structures. 

Figure\,\ref{fig:procedure} illustrates our approach. 
We start by identifying the pivoting token in a transcription we want to augment, here ``playing'' in the sentence ``two children are \textit{playing} on a statue''. Then, we extract the list of possible suffixes following ``playing'' from the suffix memory and sample a single audio-text suffix, here ``volleyball in a park''. Together with the original prefix and pivoting token, the textual part of the sampled suffix builds a new augmented transcription. Similarly, together with the audio prefix and token, the audio part of the suffix builds a new augmented audio example. The augmented transcription is then translated by an MT model. The new audio example (i.e., the representation of ``two children playing volleyball in a park'') and the translation (i.e., the text ``Zwei Kinder spielen Volleyball in einem Park'') are then recombined to form a new audio-translation pair.

\section{Experimental Setting}

\paragraph{Data Preprocessing} 
We evaluate our method on two common AST datasets, CoVoST\,2 \cite{wang2020covost} and Europarl-ST \cite{iranzo2020europarl}. 
Since Europarl-ST is too small for MT training from scratch, we use 1.6M En-De sentence pairs from Wikipedia following  \citet{schwenk-etal-2021-wikimatrix} and 3.2M En-Fr sentence pairs from the Common Crawl corpus\footnote{\href{https://www.statmt.org/wmt13/translation-task.html}{www.statmt.org/wmt13/...}, accessed 3/11/2022} as additional data. More details on the datasets are in Appendix \ref{sec:datadesc}.

For speech data preprocessing, we extract log Mel-filter banks of 80 dimensions computed every 10ms with a 25ms window. We normalize the speech features per channel using mean and variance per instance. 
For all textual data, punctuation is normalized using \textsc{sacremoses}.\footnote{\href{https://github.com/alvations/sacremoses}{github.com/alvations/sacremoses}, accessed 3/11/2022} The transcriptions are lowercased with punctuation removed. 

For the speech-to-text tasks on CoVoST\,2, we employ character-level models due to the availability of pre-trained high quality ASR models.
For the speech-to-text tasks on Europarl-ST, we learn 5,000 subword units for each target language. 
For the machine translation tasks in knowledge distillation, we learn a joint subword vocabulary on both source and target for each language pair of size 5,000 for CoVoST\,2 and size 40,000 for Europarl-ST including the additional training data.
Subword unit creation is always conducted with \textsc{sentencepiece} \cite{kudo2018sentencepiece}.

The Montreal Forced Aligner \cite{mcauliffe2017montreal} is applied without any fine-tuning to extract the acoustic alignments. Thus, the obtained alignments can be of low quality
and we discard such examples from our augmentation procedure. Please refer to Appendix \ref{sec:filtercrit} for details on our filtering criteria.
To extract POS-tags, we use the \textsc{spaCy}\footnote{\href{https://github.com/explosion/spaCy}{github.com/explosion/spaCy}, accessed 3/11/2022}
toolkit. We select the verb as our pivoting token and generate the suffix memory as follows: for each verb, we generate a list of audio-text suffix pairs and store the data in a key-value table. The audio entries contain only references to the original audio segments and our implementation is thus very memory efficient. We only utilize basic off-the-shelf components that are widely available and our suffix memory has a negligible memory footprint.
Table \ref{tab:data} summarizes the number of additional training examples in each experiment.

\begin{table}[h]
\centering
\footnotesize
{
\begin{tabular}{lccc}
\toprule
Data & Baseline & KD & STR \\
\midrule
CoVoST\,2 & $288$k & $+288$k & $+255$k  \\
Europarl-ST (En-De) & $3.25$k  & $+3.25$k & $+2.78$k \\
Europarl-ST (En-Fr) & $3.17$k & $+3.17$k & $+2.71$k  \\
\bottomrule
\end{tabular}
}
\caption{Number of examples per configuration.
}
\label{tab:data}
\vspace{-2mm}
\end{table}

\paragraph{Model configuration}
All our implementations are based on \textsc{fairseq} \cite{wang2020fairseqs2t,ott2019fairseq}.\footnote{\href{https://github.com/statnlp/str/}{github.com/statnlp/str/}, accessed 3/10/2022} In all speech-to-text tasks, we use the Transformer architecture \cite{vaswani2017attention} labelled as ``s2t\_transformer\_s'' in \textsc{fairseq}, which consists of convolutional layers for downsampling the input sequence with a factor of 4 before the self-attention layers. 
The encoder has 12 layers while the decoder has 6 layers with the dimensions of the self-attention layers set to 256 and the feed-forward network dimension set to 2048. 

For the CoVoST\,2 MT tasks, we use a smaller Transformer model of 3 layers for both encoder and decoder. The encoder-decoder embeddings and the output layer are shared. For the Europarl-ST MT tasks, we use the Transformer BASE configuration as described in \citet{vaswani2017attention}. 

\paragraph{Training} 
In the CoVoST\,2 AST experiments, we use the character-level ASR model downloaded from the \textsc{fairseq} GitHub webpage
\footnote{\href{https://github.com/pytorch/fairseq/blob/main/examples/speech_to_text/docs/covost_example.md}{github.com/pytorch/fairseq/...}, accessed 3/11/2022}
to initialize the encoder of the AST systems. Each AST system is then trained for another 50,000 steps. For Europarl-ST, we train a subword unit ASR system on the English audio-transcription pairs of the En-De data for 25,000 steps. The resulting ASR system is used to initialize both En-De and En-Fr AST systems which are trained for another 20,000 steps. 
Throughout all speech-to-text experiments, we apply gradient accumulation resulting in an effective mini-batch size of 160k frames. We use Adam optimizer \cite{kingma-ba-2014} with an inverse square root learning rate schedule. We use 10k steps for warmup and a peak learning rate of 2e-3. SpecAugment \cite{Park2019SpecAugmentAS} is applied with a frequency mask parameter of 27 and a time mask parameter of 100, both with 1 mask along their respective dimension. We perform validation and checkpoint saving after every 1,000 updates. 

In case of the CoVoST\,2 MT task, the Transformer model is pre-trained on in-domain data with 30,000 steps and an effective mini-batch size of 16,000 tokens. For the Europarl-ST dataset, the MT models are pre-trained on a combination of Europarl-ST and the additional training data. 
The Adam optimizer is used with an inverse square root learning rate schedule again, now with 4k steps for warmup and a peak learning rate of 5e-4. After pre-training, we finetune the model on the in-domain data with SGD and a constant learning rate of 5e-5.

\paragraph{Inference} In the speech-to-text experiments, we average the 10 best checkpoints based on the validation loss. For the MT tasks, we average the 5 best checkpoints. Throughout all AST experiments and MT tasks, we apply beam search with a beam size of 5.

\begin{table*}[ht]
\centering
\footnotesize
{
\begin{tabular}{lcccccc}
\toprule
model & En-De & En-Ca & En-Tr & En-Cy & En-Sl \\
\midrule
\citet{wang2020covost} Bi-AST & 16.3 & 21.8 & 10.0 & 23.9 & 16.0 \\
Baseline & $17.22$ $\pm$ $0.09$ & $23.15$ $\pm$ $0.10$ & $10.31$ $\pm$ $0.04$ & $25.46$ $\pm$ $0.08$ & $15.64$ $\pm$ $0.04$ \\
KD & $18.26$ $\pm$ $0.05$  & $24.48$ $\pm$ $0.16$ & $11.10$ $\pm$ $0.03$  & $26.87$ $\pm$ $0.16$ & $17.21$ $\pm$ $0.02$ \\
STR & $18.77$ $\pm$ $0.04$ & $24.83$ $\pm$ $0.12$ & $11.62$ $\pm$ $0.04$ & $27.28$ $\pm$ $0.11$ & $17.54$ $\pm$ $0.14$ \\
KD+STR & $19.06$ $\pm$ $0.02$ & $25.33$ $\pm$ $0.06$  & $11.83$ $\pm$ $0.01 $ & $27.73$ $\pm$ $0.09$ & $17.83$ $\pm$ $0.09$ \\
\bottomrule
\end{tabular}
}
\caption{Averaged results in BLEU on the CoVoST\,2 dataset over 3 runs with standard deviations ($\pm$). Models KD and KD+STR are significantly different for all language pairs with $p$ < 0.0002 using a paired randomization test.
}
\label{tab:covost:bleu}
\end{table*}

\section{Results}

Our experiments are focused on the improvements of our proposed method over KD alone on both CoVoST\,2 and Europarl-ST datasets. We evaluate the translation results with both BLEU\footnote{nrefs:1|case:mixed|eff:no|tok:13a|smooth:exp|version:2.0.0}
\cite{papineni-etal-2002-bleu} and chrF2\footnote{nrefs:1|case:mixed|eff:yes|nc:6|nw:0|space:no|version:2.0.0}
\cite{popovic-2016-chrf} using the implementation of \textsc{sacrebleu} \cite{post-2018-call}. Each experiment is repeated 3 times and we report mean and standard deviation. 

We also conduct significance tests using a paired approximate randomization test \cite{riezler2005some} with default settings of \textsc{sacrebleu}. 
We compute $p$-values between KD and KD+STR for each evaluated language pair 
of the experiments' datasets and between each run initialized with the same random seed.
The individual $p$-values are reported in Appendix \ref{sec:pvaluesdetail}.

Section \ref{sec:dependmtperf} contains a discussion on the connection between STR- and MT-performance. 
We also report additional experiments which show how the amount of STR data affects the final performance in Section \ref{sec:dependamount}. 
An error analysis with examples and a discussion on the limitations of STR has been moved to Appendix \ref{sec:exampleserror} due to space constraints.

\subsection{Results on CoVoST\,2}

Table \ref{tab:covost:bleu} lists BLEU scores on the five considered CoVoST\,2 language pairs. Our baseline model is the AST system finetuned on the in-domain audio-translation pairs only. Its performance over the selected language pairs is quite diverse with BLEU scores ranging from 10.31 (En-Tr) to 25.46 (En-Cy). Our baseline models are comparable to and often better in terms of BLEU than the bilingual AST (Bi-AST) models by \citet{wang2020covost}. 

Training together with translations generated by KD improves the baseline model by a substantial margin of 0.8 to 1.6 BLEU points.
Our proposed STR method alone slightly surpasses the KD performance and brings further improvements when the augmented data is combined (KD+STR) with BLEU score increases ranging from 0.62 for En-Sl to 0.86 for En-Cy. In total, we observe BLEU score improvements of 1.5 to 2.3 for KD+STR.
Since BLEU scores are often biased towards short translations, we additionally calculate chrF2 scores and report them in Appendix \ref{sec:addchrf2scrores}.

We obtain significantly different models for all language pairs with $p$ < 0.0002. This is strong evidence that the better performing models trained on KD+STR are different to the plain KD models.

\subsection{Results on Europarl-ST}

Table 3 lists the BLEU score results of Europarl-ST En-De and En-Fr. 
Similar to the results on CoVoST\,2, the KD models bring substantial improvements over the baseline systems. The gains are 6.02 points for En-De and 6.27 points for En-Fr. We attribute this to the strong machine translation model
that is trained on large amounts of additional training data (see Section \ref{sec:dependmtperf} for more details on this).
Our proposed STR method alone does not reach the KD performance but the combination KD+STR still delivers remarkable gains over KD, i.e., 1.13 points on En-De and 0.45 points on En-Fr, showing the complementarity of KD and STR. We also evaluate our models using chrF2. The numbers are listed in Appendix \ref{sec:addchrf2scrores}.

\begin{table}[ht]
\centering
\footnotesize
{
\begin{tabular}{lcc}
\toprule
model & En-De & En-Fr \\
\midrule
Baseline & $14.47$ $\pm$ $0.16$ & $22.52$ $\pm$ $0.07$  \\
KD & $20.49$ $\pm$ $0.07$ & $28.79$ $\pm$ $0.14$ \\
STR & $19.80$ $\pm$ $0.14$ & $28.01$ $\pm$ $0.17$ \\
KD+STR & $21.62$ $\pm$ $0.12$ & $29.28$ $\pm$ $0.10$   \\
\bottomrule
\end{tabular}
}
\caption{Averaged results in BLEU on the Europarl-ST dataset over 3 runs with standard deviations ($\pm$). Models KD and KD+STR are significantly different for En-De with $p$ < 0.00025. 
For En-Fr, we only found two runs to be significantly different with $p$ < 0.05. 
}
\label{tab:europarl:bleu}
\end{table}

In the En-De experiments, we obtain significant differences between the KD and KD+STR models with  $p$ < 0.00025. 
For En-Fr, only two out of three runs show significant differences with $p$ < 0.05. In terms of chrF2 scores, however, we found all compared models to be significantly different. See Appendix \ref{sec:pvaluesdetail} for details.

\subsection{Connection to MT-Performance}
\label{sec:dependmtperf}

To evaluate the dependency of STR on the MT-performance, we calculate BLEU scores for the MT-systems we use for CoVoST\,2 and Europarl-ST data augmentation with STR and compare them in a cross-lingual manner. We see a noticeable correlation of MT-performance and STR-improvement. 

On CoVoST\,2, the highest improvement for STR is observed on  the En-Cy language pair, which is also the best performing MT-model. The En-Ca language pair's MT-model also performs very well and shows the second highest gain for STR together with En-Sl. See Table \ref{tab:covost-MT} for more details.

On Europarl-ST, we observe a different behavior. While the MT-model for En-Fr is clearly better than the one for En-De, the gains are larger in the latter case. This might be due to the fact that the En-Fr ST-model already has a relatively high performance after training on KD alone (see Table \ref{tab:europarl:bleu}). We also hypothesize that adding our STR method to KD is more useful if the sentence structure of source and target languages is very different. In case the alignments between source and target language are relatively parallel, KD already generates very useful examples and our approach can only introduce limited new information on top of that, e.g., by adding speaker variations.
See Table \ref{tab:europarl-MT} for the exact BLEU scores and improvements. 

\begin{table}[ht]
\centering
\footnotesize
{
\begin{tabular}{lccccc}
\toprule
model & En-De & En-Ca & En-Tr & En-Cy & En-Sl \\
\midrule
MT & $30.05$ & $39.66$ & $21.28$ & $43.57$ & $30.32$ \\
STR-$\Delta$ & $+1.84$ & $+2.18$ & $+1.51$& $+2.27$& $+2.19$ \\
\bottomrule
\end{tabular}
}
\caption{Machine translation performance measured in BLEU on the CoVoST\,2 test set. The second row (STR-$\Delta$) reports the BLEU improvements of KD+STR in comparison to the baseline. 
}
\label{tab:covost-MT}
\end{table}

\begin{table}[ht]
\centering
\footnotesize
{
\begin{tabular}{lcc}
\toprule
model & En-De & En-Fr\\
\midrule
MT & $32.16$ & $40.11$ \\
STR-$\Delta$ & $+7.15$ & $+6.76$ \\
\bottomrule
\end{tabular}
}
\caption{Machine translation performance measured in BLEU on the Europarl-ST test set. The second row (STR-$\Delta$) reports  BLEU improvements of KD+STR in comparison to the baseline.
}
\label{tab:europarl-MT}
\end{table}

\subsection{Dependence on Amount of STR Data}
\label{sec:dependamount}

\begin{figure}
\centering
\includegraphics[width=1.05\linewidth]{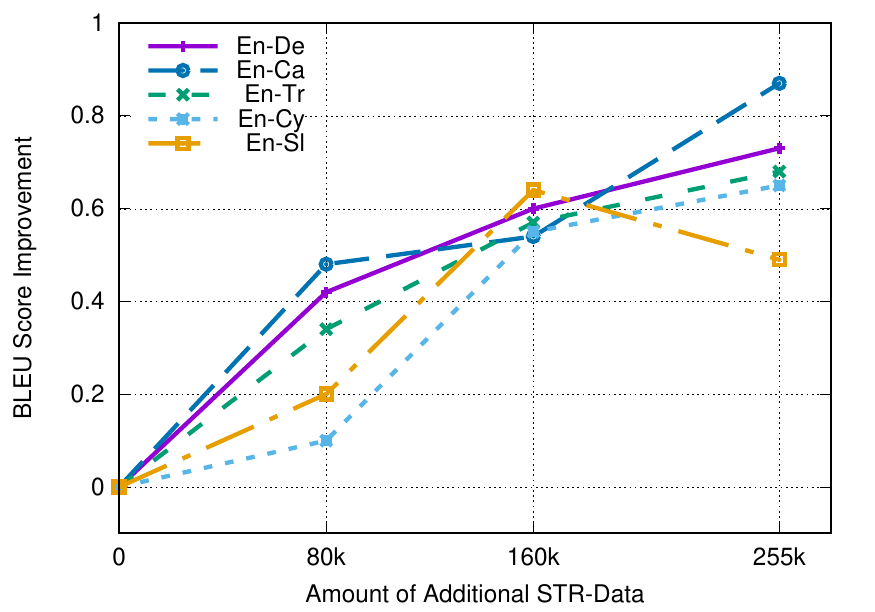}
\caption{BLEU improvements for different amounts of STR augmented data on CoVoST\,2 on a single run (seed=0) for 5 language pairs. We evaluate the addition of 0, 80k, 160k, and 255k STR-generated data points to the baseline KD data.}
\label{fig:bleuvsamount}
\end{figure}

We conduct an additional experiment on CoVoST\,2 to evaluate the dependence of our STR method on the amount of generated training data. 
In Figure \ref{fig:bleuvsamount} we report the test performance on 5 language pairs of a single run (seed=0) after training on 1/3, 2/3, or all STR generated data. For some language pairs, we already observe large gains after using 1/3 or 2/3 of the total STR data. Most language pairs will further benefit from more additional data, while one language pair (En-Sl) seems to degrade when moving from 2/3 to all training data on this single run. 
Summarizing, we observe a trend on all but one language pair that more augmented data improves performance. 

\section{Conclusion}

We proposed an effective data augmentation method for end-to-end speech translation which 
leverages audio alignments, linguistic properties, and translation. It creates new audio-translation pairs via \textit{sampling} from a memory-efficient suffix memory, \textit{translating} through an MT model and \textit{recombining} original and sampled audio segments with translations. Our method achieves significant improvements over augmentation with KD alone on both large (CoVoST\,2) and small scale (Europarl-ST) datasets. In future work, we would like to investigate the utility of other linguistic properties for AST augmentation and we would like to extend our method to multilingual AST.

\section*{Acknowledgements}

This research was supported in part by the German research foundation DFG under grant RI-2221/4-1.
We'd also like to thank the anonymous reviewers for their helpful comments.

\bibliographystyle{acl_natbib}
\bibliography{references}

\appendix
\section{Appendix}
\label{sec:appendix}

\subsection{Data Description}
\label{sec:datadesc}
CoVoST\,2 is a large scale dataset of 430h English audio and 288k sentences for each language in the training set. The training set contains repetitions of the same sentence spoken by different speakers. We use the original data splits generated by the \texttt{get\_covost\_splits.py} script\footnote{\href{https://github.com/facebookresearch/covost}{github.com/facebookresearch/covost}, accessed 3/11/2022} on five languages pairs, namely English-German (En-De), English-Catalan (En-Ca), English-Turkish (En-Tr), English-Welsh (En-Cy) and English-Slovenian (En-Sl), resulting in about 15.5k sentences for each dev and test dataset. 

Europarl-ST, in contrast, is a small AST dataset. 
It contains debates held in the European Parliament and their translations, thus representing a realistic AST scenario imposing very different challenges than the CoVoST\,2 dataset. 
We conduct experiments on the English-German (En-De) and English-French (En-Fr) language pairs. The En-De data contains 89h of audio and 35.5k sentences. The En-Fr data contains 87h of audio and 34.5k sentences.

\subsection{Filtering Criteria by the Acoustic Aligner}
\label{sec:filtercrit}

In very rare cases, the acoustic aligner does not return an alignment at all and we have to discard these examples. In some cases, the obtained alignments by the acoustic aligner are of low quality, i.e., contain alignments to unknown tokens. In such cases, if the number of tokens of the output transcriptions of the acoustic aligner matches the number of tokens in the input transcriptions, we can still use this alignment for data augmentation as alignments in ASR are always strictly parallel. Thus, if we cannot retrieve suitable alignments, we discard the example. This procedure reduces the amount of augmented data: we discard approximately 12\% of the examples for CoVoST\,2, and about 15\% of the examples for Europarl-ST. See Table \ref{tab:data} for the final data sizes.

\subsection{Additional chrF2 Scores}
\label{sec:addchrf2scrores}
In this section, we additionally report chrF2 scores on CoVoST\,2 and Europarl-ST datasets,
since BLEU scores are often biased towards short translations. This issue is especially problematic on the CoVoST\,2 datasets because of its large number of very short sentences. 
We list the CoVoST\,2 chrF2 results in Table\,\ref{tab:covost:chrF}, and the Europarl-ST results in  Table\,\ref{tab:europarl:chrf}.

Our chrF2 results averaged over three runs confirm the improvements we observed throughout our experiments in terms of BLEU. When we look at chrF2, the better performing KD+STR models are always significantly different to the KD models. Even in case of the En-Fr language pair of the Europarl-ST dataset where we detected significant differences only in two of three runs in terms of BLEU, we found all three runs significantly different in terms of chrF2 with $p$ < 0.025 this time. 
Detailed $p$-values per run are listed in Table\,\ref{tab:covost-chrF-p} and \ref{tab:covost-bleu-p} for our CoVoST\,2 experiments, and in Table\,\ref{tab:europarl-p} for our Europarl-ST experiments.

\begin{table}[h]
\centering
\footnotesize
{
\begin{tabular}{lcc}
\toprule
model & En-De & En-Fr \\
\midrule
Baseline & $44.90$ $\pm$ $0.22$ & $48.60$ $\pm$ $0.14$  \\
KD & $51.43$ $\pm$ $0.05$  & $54.97$ $\pm$ $0.05$ \\
STR & $50.6$ $\pm$ $0.0$ &  $54.1$ $\pm$ $0.22$ \\
KD+STR & $52.37$ $\pm$ $0.09$ & $55.37$ $\pm$ $0.09$ \\
\bottomrule
\end{tabular}
}
\caption{Averaged results in chrF2 on En-De and En-Fr of Europarl-ST dataset over 3 runs with standard deviations ($\pm$). Models KD and KD+STR are significantly different for En-De with $p$ < 0.0002 using a paired randomization test. For En-Fr, the models are significantly different with $p$ < 0.025.
}
\label{tab:europarl:chrf}
\end{table}

\subsection{Detailed $p$-values for System Comparison}
\label{sec:pvaluesdetail}

Tables \ref{tab:covost-bleu-p} and \ref{tab:covost-chrF-p} report the exact $p$-values for comparison of KD and KD+STR models w.r.t. BLEU and chrF2 scores on CoVoST\,2, respectively.
Table \ref{tab:europarl-p} reports the exact $p$-values 
for comparison of KD and KD+STR models w.r.t. BLEU and chrF2 scores on Europarl-ST, respectively. We use the implementation of \textsc{sacrebleu} for calculation.

\begin{table}[ht]
\centering
\footnotesize
{
\begin{tabular}{lcccccc}
\toprule
seed & En-De & En-Ca & En-Tr & En-Cy & En-Sl \\
\midrule
0 & $0.0001$ & $0.0001$ & $0.0001$ & $0.0001$ & $0.0001$ \\
1 & $0.0001$ & $0.0001$ & $0.0001$ & $0.0001$ & $0.0001$ \\
321 & $0.0001$ & $0.0001$ & $0.0001$ & $0.0001$ & $0.0001$ \\
\bottomrule
\end{tabular}
}
\caption{The paired randomization test from \textsc{sacrebleu} with default settings returned the following $p$-values for the three runs when comparing KD and KD+STR models' BLEU performance on CoVoST\,2.
}
\label{tab:covost-bleu-p}
\end{table}

\begin{table}[ht]
\centering
\footnotesize
{
\begin{tabular}{lcccccc}
\toprule
seed & En-De & En-Ca & En-Tr & En-Cy & En-Sl \\
\midrule
0 & $0.0001$ & $0.0001$ & $0.0001$ & $0.0001$ & $0.0001$ \\
1 & $0.0001$ & $0.0001$ & $0.0001$ & $0.0001$ & $0.0001$ \\
321 & $0.0001$ & $0.0001$ & $0.0001$ & $0.0001$ & $0.0001$ \\
\bottomrule
\end{tabular}
}
\caption{The paired randomization test from \textsc{sacrebleu} with default settings returned the following $p$-values for the three runs when comparing KD and KD+STR models' chrF2 performance on CoVoST\,2.
}
\label{tab:covost-chrF-p}
\end{table}

\begin{table}[h]
\centering
\footnotesize
{
\begin{tabular}{lcccc}
\toprule
     & \multicolumn{2}{c}{BLEU} & \multicolumn{2}{c}{chrF2} \\
seed & En-De & En-Fr & En-De & En-Fr \\
\midrule
0 & $0.0002$ & $0.010$ & $0.0001$ & $0.016$  \\
1 & $0.0001$ & $0.137$ & $0.0001$ & $0.021$ \\
321 & $0.0002$ & $0.012$ & $0.0001$ & $0.005$ \\
\bottomrule
\end{tabular}
}
\caption{Conducting the paired randomization test from \textsc{sacrebleu} with default settings returned the following $p$-values for the three runs when comparing KD and KD+STR models' performance on Europarl-ST. 
}
\label{tab:europarl-p}
\end{table}

\subsection{Examples and Error Analysis}
\label{sec:exampleserror}

We also take a look at the quality of our STR-augmented data and list examples in Table \ref{tab:covost-examples} and Table \ref{tab:europarl-examples} for CoVoST\,2 and Europarl-ST, respectively. 
Rows ``src-A'' and ``src-B'' contain the unmodified transcriptions from CoVoST\,2 with our pivoting token underlined and segments we recombine in \textit{italics}. The row ``augm.'' shows the STR-augmented example, the row ``transl.'' contains the MT-generated translation.
The presented examples are the first 5 data examples taken directly from our augmented data set and are \textit{not} cherry-picked. 

Of the first five augmented examples from CoVoST\,2 listed in Table \ref{tab:covost-examples}, examples 1, 3, and 5 contain grammatically correct augmented source data (row ``augm.'') and the latter two are also semantically correct. 
Example 2 contains a grammatically wrong segment due to the problematic transcription of ``src-B'': here, the example is already an ungrammatical sentence and this transfers to our augmented example. Example 4 is also grammatically wrong. In this example, our augmentation method mixes the different senses of the word ``directed'' and produces a semantically incorrect result. This could be fixed by integrating more context, e.g., ``directed through'' can be used to disambiguate the different word senses of ``directed''.

Of the first five augmented examples from Europarl-ST in Table \ref{tab:europarl-examples}, examples 1, 3, and 5 are actually grammatically correct. 
Example 2 is grammatically wrong as our STR method does not respect the different grammatical forms of ``pass'' in ``will pass'' and ``to pass'', mixing up the two objects. 
Example 4 is also grammatically wrong, and it is again the wrong treatment of different grammatical forms of ``do'' in ``do work'' and ``to do''. These problems could be addressed by putting more effort into the suffix memory construction, e.g., by using n-grams as keys. 
Examples 3 and 5 demonstrate a property of Europarl-ST that partly explains the lower performance gain we observe for our STR-method here: there are many repetitive formalized sentences, and in these examples our augmentation method only differs by a single word from an already existing data example. Still, such augmented examples can be useful for training due to the speaker variations injected by STR.

We observe common errors in our augmented examples for CoVoST\,2 and Europarl-ST that are often connected to the different word senses and syntactical functions of the selected pivoting token. However, even grammatically wrong sentences can sometimes be useful in training as they prevent overfitting on common structures in the data. Furthermore, the speaker variations in the examples that we produce can be helpful even if the augmented examples do not differ much from existing ones. Summarizing the error analysis, our simple STR-method is able to produce examples that are useful even with errors. Investigating more complex methods for better identification of pivoting tokens is a promising direction for future work.

\begin{table*}[ht]
\centering
\footnotesize
{
\begin{tabular}{lcccccc}
\toprule
model & En-De & En-Ca & En-Tr & En-Cy & En-Sl \\
\midrule
Baseline & $42.80$ $\pm$ $0.08$ & $46.63$ $\pm$ $0.09$ & $36.77$ $\pm$ $0.09$ & $49.13$ $\pm$ $0.05$ & $39.83$ $\pm$ $0.05$ \\
KD & $44.13$ $\pm$ $0.05$ & $48.17$ $\pm$ $0.12$  & $38.53$ $\pm$ $0.05$ & $50.67$ $\pm$ $0.05$ & $41.73$ $\pm$ $0.05$  \\
STR & $44.43$ $\pm$ $0.05$ & $48.60$ $\pm$ $0.08$ & $39.30$ $\pm$ $0.08$ & $51.03$ $\pm$ $0.05$ & $42.17$ $\pm$ $0.05$ \\
KD+STR &  $45.13$ $\pm$ $0.05$ & $49.10$ $\pm$ $0.08$  & $39.70$ $\pm$ $0.08$ & $51.50$ $\pm$ $0.00$ & $42.60$ $\pm$ $0.08$ \\
\bottomrule
\end{tabular}
}
\caption{Averaged results in chrF2 on the CoVoST\,2 dataset over 3 runs with standard deviations ($\pm$). Models KD and KD+STR are significantly different for all language pairs with $p$ < 0.0002 using a paired randomization test.
}
\label{tab:covost:chrF}
\end{table*}

\begin{table*}[ht]
\centering
\footnotesize
{
\begin{tabular}{lll}
\toprule
& src-A & \textit{these data components in turn} \underline{serve} as the building blocks of data exchanges\\
& src-B & the governor appoints members of the board each of whom \underline{serve} \textit{seven years}\\
1& augm. & \textit{these data components in turn \underline{serve} seven years}\\
& transl. & Diese Datenkomponenten wiederum servieren sieben Jahre.\\
\midrule
& src-A & \textit{the church} \underline{is} unrelated to the jewish political movement of zionism\\
& src-B & both sacks contain a man b \underline{is} \textit{on the left a on the right}\\
2& augm. & \textit{the church \underline{is} on the left a on the right}\\
& transl. & Die Kirche befindet sich rechts auf der linken Seite.\\
\midrule
& src-A & \textit{the following} \underline{represents} architectures which have been utilized at one point or another\\
& src-B & monism sees brahma as the ultimate reality while monotheism \underline{represents} \textit{the personal form brahman}\\
3& augm. & \textit{the following \underline{represents} the personal form brahman}\\
& transl. & Die folgende Darstellung repräsentiert die persönliche Form Brahman.\\
\midrule
& src-A & \textit{additionally the pulse output can be} \underline{directed} through one of three resonator banks\\
& src-B & he \underline{directed} \textit{no fewer than thirty seven productions at stratford}\\
4& augm. & \textit{additionally the pulse output can be \underline{directed} no fewer than thirty seven productions at stratford}\\
& transl. & Darüber hinaus kann der Pulsausgang nicht weniger als siebenunddreißig Produktionen in Stratford geleitet \\
& & werden.\\
\midrule
& src-A & \textit{the two} \underline{are} robbed by a pickpocket who is losing in gambling\\
& src-B & there \underline{are} \textit{six large portraits displayed in the senate chamber}\\
5& augm. & \textit{the two \underline{are} six large portraits displayed in the senate chamber}\\
& transl. & Die beiden sind sechs große Porträts, die in der Senatskammer ausgestellt sind.\\
\bottomrule
\end{tabular}
}
\caption{The first 5 augmented data examples from CoVoST\,2 for the En-De language pair. ``src-A'' and ``src-B'' are the unmodified transcriptions from CoVoST\,2 with our pivoting token underlined and segments we recombine in \textit{italics}. The ``augm.'' row shows the STR-augmented example. The ``transl.'' row contains the MT-generated translation.
}
\label{tab:covost-examples}
\end{table*}

\begin{table*}[ht]
\centering
\footnotesize
{
\begin{tabular}{lll}
\toprule
& src-A & \textit{i would just like to say that there are more amendments in my report because my committee} \underline{has} been more \\
&  & ambitious in the improvements it wanted to make to the commission proposal\\
& src-B & economic cooperation \underline{has} \textit{always been europe s most powerful engine for greater integration and europe has } \\
&  & \textit{owed its success to this pragmatic approach since 1956}\\
1& augm. & \textit{i would just like to say that there are more amendments in my report because my committee \underline{has} always been }\\
&  & \textit{europe s most powerful engine for greater integration and europe has owed its success to this pragmatic } \\
&  & \textit{approach since 1956}\\
& transl. & Je voudrais juste dire qu ' il y a plus de modifications dans mon rapport, car ma commission a toujours été le \\
&  & moteur le plus puissant de l ' Europe pour une plus grande intégration, et l ' Europe doit son succès à cette  \\
&  & approche pragmatique depuis 1956.\\
\midrule
& src-A & \textit{i would like to thank all my colleagues on the committee who worked with me to put together some really big} \\
&  & \textit{compromise amendments which we will} \underline{pass} today\\
& src-B & the right of every member state to \underline{pass} \textit{laws as it deems fit as long as it has a democratic majority and that} \\
&  & \textit{those laws should be recognised by other countries}\\
2& augm. & \textit{i would like to thank all my colleagues on the committee who worked with me to put together some really big} \\
&  & \textit{compromise amendments which we will \underline{pass} laws as it deems fit as long as it has a democratic majority and } \\
&  & \textit{that those laws should be recognised by other countries}\\
& transl. & Je tiens à remercier tous mes collègues de la commission qui ont travaillé avec moi pour mettre en place des \\ 
&  & amendements de compromis vraiment importants, que nous adopterons des lois, tant qu ' elle a une majorité \\
&  & démocratique et que ces lois devraient être reconnues par d ' autres pays.\\
\midrule
& src-A & \textit{i would} \underline{like} all of you to give us a huge majority for this so that when we come to negotiate with the  \\
&  & commission and council we will do our very best for europe s consumers\\
& src-B & i would also \underline{like} \textit{to thank all the shadow rapporteurs}\\
3& augm. & \textit{i would \underline{like} to thank all the shadow rapporteurs}\\
& transl. & Je tiens à remercier tous les rapporteurs fictifs.\\
\midrule
& src-A & \textit{mr president let us hope that the american proposals for purchases of toxic assets} \underline{do} work because if they do\\
&  & not the contagion will almost certainly spread over here\\
& src-B & what we really need to \underline{do} \textit{is empower women} \\
4& augm. & \textit{mr president let us hope that the american proposals for purchases of toxic assets \underline{do} is empower women} \\
& transl. & Monsieur le Président, espérons que les propositions américaines d ' achats d ' actifs toxiques permettent aux \\
&  & femmes.\\
\midrule
& src-A & \textit{i would} \underline{like} assurance from mr jouyet and mr almunia that we really do have our defences in place\\
& src-B & mr president i would \underline{like} \textit{to thank the rapporteurs and other shadows for the hard work they have put into} \\
&  & \textit{producing these reports}\\
5& augm. & \textit{i would \underline{like} to thank the rapporteurs and other shadows for the hard work they have put into producing these} \\
&  & \textit{reports}\\
& transl. & Je voudrais remercier les rapporteurs et d ' autres ombres pour le travail qu ' ils ont accompli dans la  \\
&  & production de ces rapports.\\
\bottomrule
\end{tabular}
}
\caption{The first 5 augmented data examples from Europarl-ST for the En-Fr language pair. ``src-A'' and ``src-B'' are the unmodified transcriptions from Europarl-ST with our pivoting token underlined and segments we recombine in \textit{italics}. The ``augm.'' row shows the STR-augmented example. The ``transl.'' row contains the MT-generated translation. }
\label{tab:europarl-examples}
\end{table*}

\end{document}